\documentclass{article}
\usepackage{ijcai22}

\usepackage{times}

\usepackage{soul}
\usepackage{url}
\usepackage[hidelinks]{hyperref}
\usepackage[utf8]{inputenc}
\usepackage[small]{caption}
\usepackage{graphicx}
\usepackage{amsmath}
\usepackage{booktabs}
\usepackage{multirow}
\usepackage[table,xcdraw]{xcolor}
\usepackage{algorithm}
\usepackage{algorithmic}
\usepackage{subfigure} 

\title{Boosting Adversarial Transferability of MLP-Mixer}

\author{
Haoran Lyu\thanks{Equal contribution.}$^{,1}$\and
Yajie Wang$^{*,1}$\and
Yu-an Tan$^1$\and
Huipeng Zhou$^2$\and
Yuhang Zhao$^1$\\ \And
Quanxin Zhang\thanks{Corresponding author.}$^{,1}$
\affiliations
$^1$School of Cyberspace Science and Technology, Beijing Institute of Technology\\
$^2$School of Computer Science and Technology, Beijing Institute of Technology
\emails
\{lyuhaoran, wangyajie19, tan2008, zhouhuipeng, zhaoyuhang, zhangqx\}@bit.edu.cn
}

\begin{document}

\maketitle

\begin{abstract}

The security of models based on new architectures such as MLP-Mixer and ViTs needs to be studied urgently. However, most of the current researches are mainly aimed at the adversarial attack against ViTs, and there is still relatively little adversarial work on MLP-mixer. We propose an adversarial attack method against MLP-Mixer called Maxwell's demon Attack (MA). MA breaks the channel-mixing and token-mixing mechanism of MLP-Mixer by controlling the part input of MLP-Mixer's each Mixer layer, and disturbs MLP-Mixer to obtain the main information of images. Our method can mask the part input of the Mixer layer, avoid overfitting of the adversarial examples to the source model, and improve the transferability of cross-architecture. Extensive experimental evaluation demonstrates the effectiveness and superior performance of the proposed MA. Our method can be easily combined with existing methods and can improve the transferability by up to 38.0\% on MLP-based ResMLP. Adversarial examples produced by our method on MLP-Mixer are able to exceed the transferability of adversarial examples produced using DenseNet against CNNs. To the best of our knowledge, we are the first work to study adversarial transferability of MLP-Mixer.

\end{abstract}

\begin{figure}[t]
    \centering
    \includegraphics[width=0.9\linewidth]{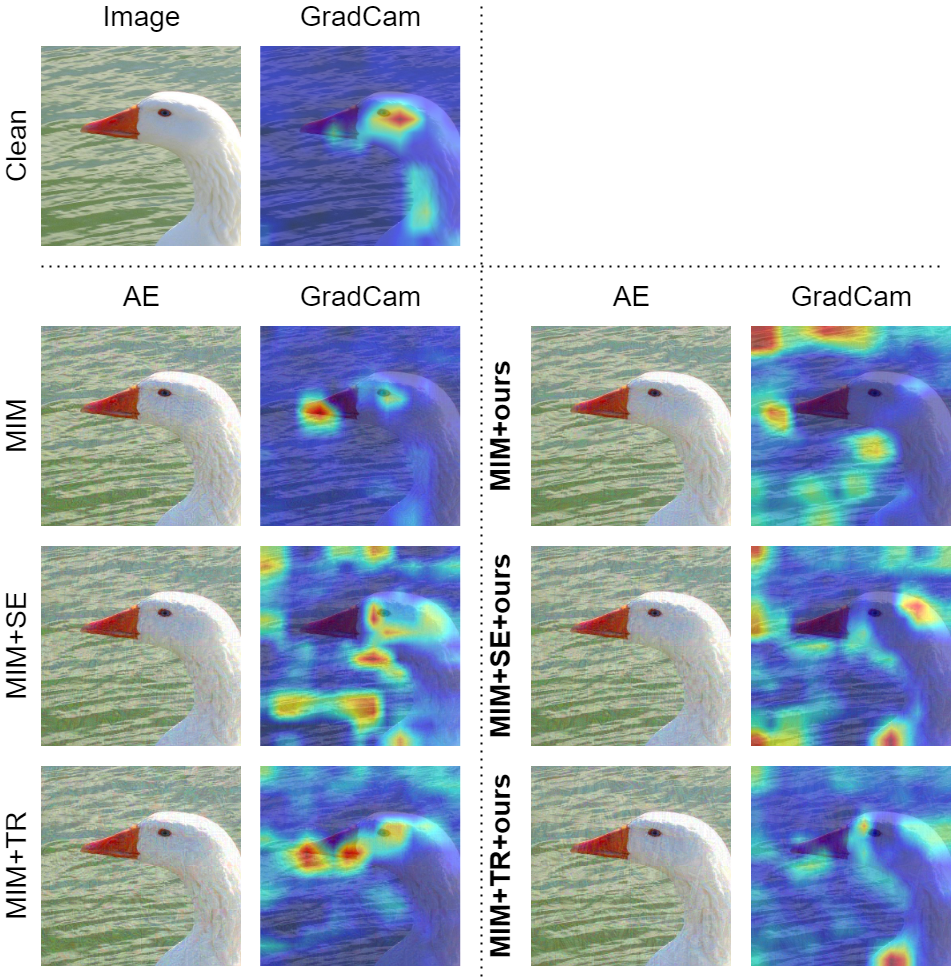}
    \caption{Demonstration of the clean image, adversarial examples (AE) made by existing adversarial attack methods and our mothods on MLP-Mixer, and their GradCam images generated on VGG-16. The adversarial examples generated by the original method still make the target model focus on the object itself. Our method can force the target model to pay more attention to the regions far from the object in the adversarial examples.}\label{fig:gradcam}
\end{figure}

\section{Introduction}

Convolutional Neural Networks (CNNs) have become the de facto standard in the field of computer vision. Deep Neural Networks (DNNs) based on CNNs continue to improve classification performance in computer vision, such as Densenet \cite{huang2017densely}, MobileNet \cite{sandler2018mobilenetv2}, EfficientNet \cite{tan2019efficientnet}, ReXNet \cite{han2021rethinking}, etc. However, with the development of attention-based transformers in the field of natural language processing, some new models applying this transformer structure have emerged, such as ViT \cite{dosovitskiy2020image}, T2T-ViT \cite{yuan2021tokens} and DeiT \cite{touvron2021training} etc. The performance of these models has caught up with CNNs and is challenging the position of CNNs in the field of computer vision. With further research, researchers found that convolution and attention mechanisms are not unique to good performance, and only using MultiLayer Perceptrons (MLPs) can also achieve good performance, so MLP-Mixer \cite{tolstikhin2021mlp} is proposed.

As we all know, DNNs have security risks and are vulnerable to adversarial examples. The adversary adds a well-designed and imperceptible perturbations to the clean input, leading DNNs to incorrect results. Due to the potential risks of DNNs, it is very important to understand whether the recently proposed ViTs and MLP-Mixer are vulnerable to adversarial attacks. The adversarial transferability of ViTs has been well studied \cite{naseer2021improving}. In contrast, MLP-Mixer has not been carefully studied in the context of black-box adversarial, and there is no research on the transferability of adversarial attacks against MLP-Mixer. In this work, we focus specifically on transfer-based adversarial attacks and study how to improve the transferability of adversarial examples generated by MLP-Mixer. 

Our analysis of MLP-Mixer is based on the following findings. MLP-Mixer differs in architecture from CNNs. Similar to ViTs, MLP-Mixer uses image patches as input, but does not use any convolution and attention mechanisms. Instead, the architecture of MLP-Mixer is entirely based on MLPs. MLP-Mixer contains two types of layers, one is mixing spatial location information, called token-mixing MLPs, and the other is mixing channel information, called channel-mixing MLPs. Information from different patches and channels can be fully mixed, enabling MLP-Mixer to capture the main information of the image. Disturbing the information mixing mechanism of MLP-Mixer can prevent the generated adversarial examples from overfitting the source models, which can improve the cross-architecture transferability of adversarial examples. 

We propose a new adversarial attack called Maxwell's demon Attack (MA). By using MA, we are able to control the input of MLP-Mixer's each Mixer layer, breaking the channel-mixing and token-mixing mechanism of MLP-Mixer. MA can mask the part input of the Mixer layer, achieve an effect similar to Dropout \cite{srivastava2014dropout}, prevent the generated adversarial examples from overfitting the MLP-Mixer, and improve the fooling rate of the adversarial examples attacking the target models.

Our proposed MA method is a detachable component that can be easily combined with existing methods. We conduct extensive experiments on models with multiple architectures using the ImageNet validation set. The adversarial examples generated by our method on MLP-Mixer attack MLP-based ResMLP can improve the fooling rate by 38.0\%, and attack ViTs can improve the fooling rate by 26.4\%. On some CNN architecture models, the adversarial examples generated by MLP-Mixer even exceed the transferability of adversarial examples generated using CNN-based DenseNet.

In summary, our main contributions are as follows:

\begin{itemize}
    \item We attack the channel-mixing and token-mixing mechanism of MLP-Mixer, which destroys the ability of MLP-Mixer to obtain the main information of images and improves the cross-architecture transferability of adversarial examples.
    
    \item We propose a new transfer-based attack called Max-well's demon Attack (MA). MA can mask the part input of MLP-Mixer's each Mixer layer, breaking the channel-mixing and token-mixing mechanism of MLP-Mixer. Our method prevents the adversarial examples from overfitting the MLP-Mixer and improves the fooling rate of attacking target models.

    \item Extensive experimental evaluation demonstrates the effectiveness and superior performance of the proposed MA,  Our method significantly improves the transferability of adversarial examples generated by MLP-Mixer, and even outperforms adversarial examples generated using DenseNet against some CNNs.

\end{itemize}

\section{Related Work}

\subsection{Adversarial Attack and Transferability}

Adversarial attacks are divided into white-box attacks and black-box attacks. The white-box attacks require access to all information about the target model, such as FGSM \cite{goodfellow2014explaining} and PGD  \cite{madry2017towards}. The black-box attacks do not need to know the target model information, and the mainstream approach is the transfer-based attack. The transfer-based attacks require an alternative model that is similar to the target model, and white-box attacks on the alternative model to generate adversarial examples. The target model is attacked by virtue of the transferability of the adversarial examples, and thus the goal of transfer-based attacks is to improve the transferability of the adversarial examples. The MIM \cite{dong2018boosting} enhances the transferability of FGSM by adding a momentum term to the gradient. The DIM \cite{xie2019improving} improves the transferability of adversarial examples by creating different input modes. The TIM \cite{dong2019evading} improves transferability by using a predefined kernel convolution on the gradient. Our method can be easily combined with these existing methods to further improve the transferability of adversarial examples.

\subsection{Robustness of The New Models}

For the robustness of the new models, we mainly focus on the robustness of ViTs and MLP-Mixer. \cite{benz2021adversarial} investigated the adversarial robustness of ViTs and MLP-Mixer. They found that MLP-Mixer is vulnerable to universal adversarial perturbations. But they did not explore the adversarial transferability of MLP-Mixer. To our knowledge, there is currently no work investigating the transferability of adversarial examples generated by MLP-Mixer. \cite{naseer2021improving} introduced two strategies to enhance the transferability of adversarial examples generated by ViTs. One is to obtain the output of each ViT block to generate adversarial examples, called Self-Ensemble, and the other is to train a classifier head for each ViT block and use the output of each classifier head to generate adversarial examples, called Token Refinement. We try to introduce these two strategies to MLP-Mixer, but the effect is not significant. After our modification and the introduction of our proposed MA, the transferability of the adversarial examples generated by MLP-Mixer is substantially improved.

\begin{figure*}[t]
  \centering
  \includegraphics[width=\textwidth]{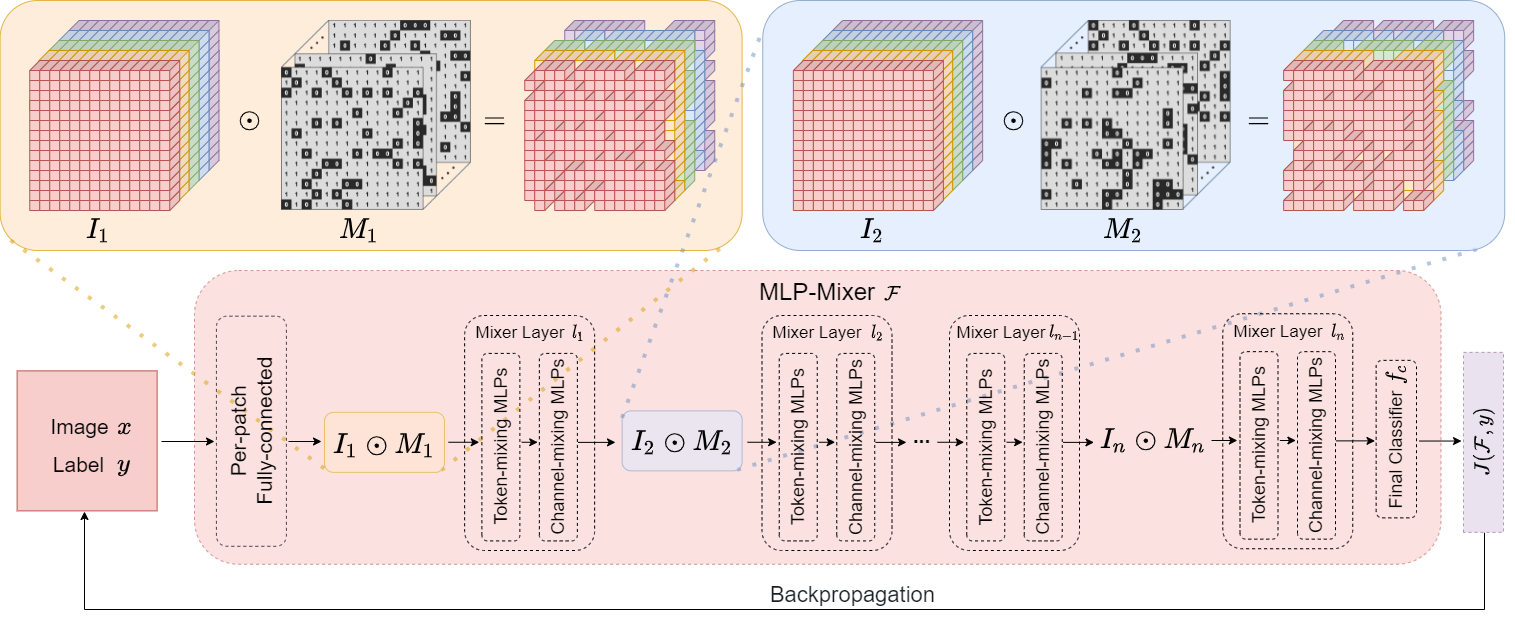}
  \caption{Maxwell's demon Attack (MA). $I_i$ represents the input of the Mixer layer $l_i$, $M_i$ is the masking matrix generated for $I_i$ based on the Bernoulli distribution, and $\odot$ represents element-wise product. After $I_i$ and $M_i$ are multiplied, some elements in $I_i$ are discarded. By masking the part input of each Mixer layer, MA breaks the token-mixing and channel-mixing of MLP-Mixer, preventing adversarial examples from overfitting MLP-Mixer, which can greatly improve the transferability of its generated adversarial examples.}
  \label{fig:ma-arch}
  \hfill
\end{figure*}

\section{Methodology}
Consider a clean image sample $x \in X$ and its ground-truth label $y \in Y$, a source model $\mathcal{F}(x): X \to Y$ and a target model $\mathcal{M}$ which is under-attack. We focus on untargeted adversarial attack, the goal of the transfer-based attack is generating an adversarial example $x_{adv}$, using the information of source model $\mathcal{F}$, which can change the target model’s prediction ($\mathcal{M}(x_{adv})\neq y$). In order to make the adversarial example imperceptible to the human eye, it is necessary to limit the modification magnitude of the adversarial example relative to the clean sample, and we use the $l_{\infty}$ for the restriction, i.e., $\Vert x_{adv}-x \Vert_{\infty} < \epsilon$. The optimization problem of generating the adversarial example is defined as follows:

\begin{equation}
    \mathop{\arg\max}\limits_{x_{adv}-x}J(\mathcal F(x_{adv}),y), s.t. \Vert x_{adv}-x \Vert_{\infty}<\epsilon
\end{equation}
where $J(\cdot,\cdot)$ is the loss function (e.g. cross-entropy).

For the MLP-Mixer model $\mathcal{F}$ with $n$ Mixer layers can be defined as:
\begin{equation}
    \mathcal F=(l_1 \circ l_2 \circ l_3 \circ \ldots \circ l_n)\circ f_c
\end{equation}
where $l_i$ represents a single Mixer layer comprising of token-mixing layer and channel-mixing layer and $f_c$ is the final classification head.

Our MA method is able to control the input of each Mixer layer. We multiply the input $I$ of each Mixer layer of MLP-Mixer by a masking matrix $M$, which can be defined as follows:

\begin{align}
    \mathcal F=(l_1(I_1 \odot M_1) \circ l_2(I_2 \odot M_2) \circ l_3(I_3 \odot M_3) \circ \ldots \nonumber\\ 
    \circ l_n(I_n\odot M_n))\circ f_c \nonumber
\end{align}
\begin{equation}
    \text{where } M_i=
    \begin{cases}
    X\sim B(1,p) & P \\
    1 & 1-P
    \end{cases} \label{eq:F}
\end{equation}
$\odot$ represents element-wise product. In the case of probability $P$, we mask the input of each Mixer layer, and $M$ is the matrix directly generated from the Bernoulli distribution.

As shown in Fig. \ref{fig:ma-arch}, our MA method controls the input of each Mixer layer, and by masking part of the input, we destroy the channel-mixing and token-mixing mechanism of MLP-Mixer, thereby improving the transferability of adversarial examples against MLP-based models. Meanwhile, by masking the input of each Mixer layer of MLP-Mixer, our method achieves a Dropout-like effect. But unlike Dropout dropping neurons, our method is to drop the feature maps of each layer, which can prevent adversarial examples from overfitting the source model MLP-Mixer, thereby improving the transferability of adversarial examples to non-MLP models, such as CNNs and ViTs.

Our method benefits from the fact that each Mixer layer structure of MLP-Mixer is the same, so it only needs to generate a masking matrix of one size, thus saving computational overhead, which is not possible in most CNNs. Our method is a detachable component that can be easily combined with existing gradient-based methods, such as PGD, MIM, DIM, TIM, etc., and the combination with the MIM algorithm is shown in Algorithm \ref{alg:algorithm}.

Our method can also be combined with the Self-Ensemble (SE) method and Token Refinement (TR) method that attack ViTs. SE and TR methods are also components that can be combined with gradient-based methods. SE obtains the output of each block of ViTs, and then inputs it into the final classifier head respectively. After obtaining all the outputs of the classifier head, SE calculates the average of the outputs as the input of the loss function. We transplant SE into MLP-Mixer and combined them with our MA. For the SE combined with MA method, it can be defined as follows:

\begin{equation}
\mathcal F_k=(\prod_{i=1}^k l_i(I_i \odot M_i))\circ f_c \text{, where } k=1,2,\ldots, n
\end{equation}
\begin{equation}
\mathcal F = \frac{1}{n}\sum_{k=1}^n F_k
\end{equation}

After masking the input of each layer of Mixer Layer, they are respectively input into the final classifier head, thus generating a Self-Ensemble of $n$ MLP-Mixer networks with different depths.

TR trains a classifier head for each block of ViTs, uses the output of each classifier head to calculate the loss value, and then averages all the loss values as the final loss value. We also transplant TR into MLP-Mixer and combined them with our MA. For the TR combined with MA method, it can be defined as follows:

\begin{equation}
\mathcal F_k=(\prod_{i=1}^k l_i(I_i \odot M_i))\circ f_c^k \text{, where } k=1,2,\ldots, n
\end{equation}
\begin{equation}
J = \frac{1}{n}\sum_{k=1}^n J_k(\mathcal F_k(x_{adv}),y)
\end{equation}
where $f_c^k$ is the classifier head we trained for each Mixer layers.

\begin{algorithm}[t]
    \caption{Maxwell's demon attack}
    \label{alg:algorithm}
    \textbf{Input}: A source MLP-Mixer model $\mathcal{F}$ composed of $n$ Mixer layers $l$ and the final classifier head $f_c$ ; a loss function $J$; a clean image $x$ with ground-truth label $y$;\\
    \textbf{Parameter}: The perturbation budget of $l_{\infty}$-normed $\epsilon$; iterations $T$; step size $\alpha$ and decay facotr $\mu$. The masking matrix $M$ generated for each Mixer layer $l$ defined by Eq. \ref{eq:F}; the input $I$ of each Mixer layer $l$\\
    \textbf{Output}: An adversarial example $x_{adv}$ with $\Vert x_{adv}-x \Vert_{\infty} \leq \epsilon$
    \begin{algorithmic}[1] 
    \renewcommand{\baselinestretch}{2}
        \STATE $x_{adv}^0 = x$; $g_0=0$
        \FOR{$t=0$ to $T-1$}
            \STATE $I_0$ = Per-Patch Fully-Connected $(x_{adv}^t)$
            \FOR{each Mixer layer $l_i$ in $\mathcal{F}$, $i=0$ to $i<n$}
                \STATE $I_{i+1}=l_i(I_i \odot M_i)$
            \ENDFOR\
            \STATE $output=f_c(I_n)$
            \STATE $g_{t+1}=\mu \cdot g_{t} + \frac{\nabla_x J(output,y)}{\Vert \nabla_x J(output,y) \Vert_1}$
            \STATE $x_{adv}^{t+1}=clip_{x,\epsilon}(x_{adv}^t+\alpha \cdot sign(g_{t+1})$
        \ENDFOR
        \STATE \textbf{return} $x_{adv}=x_{adv}^{T}$
    \end{algorithmic}
\end{algorithm}

\begin{figure}[t]
    \centering
    \includegraphics[width=0.9\linewidth]{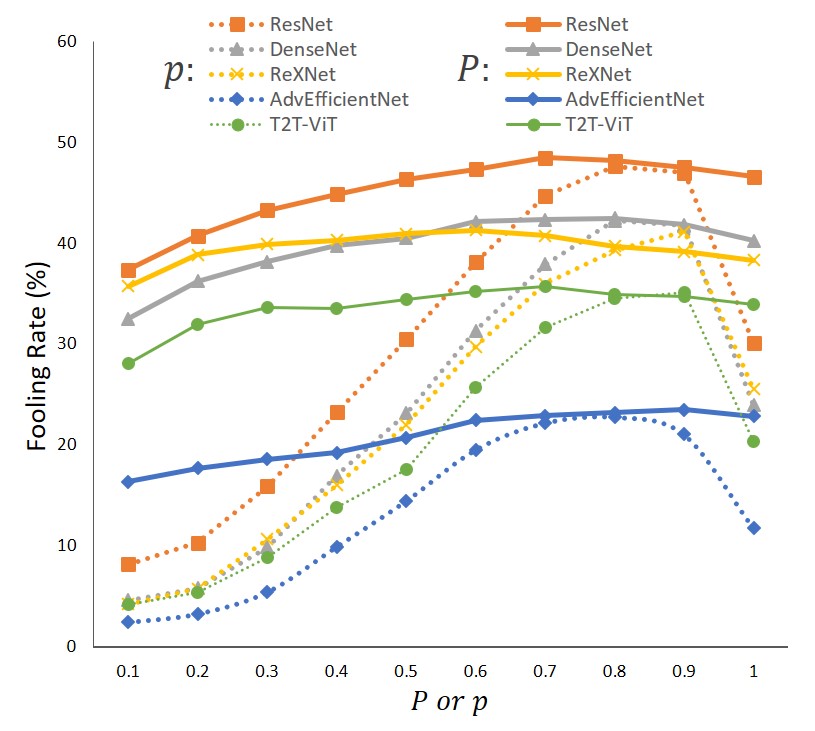}
    \caption{The experimental results of selecting the probability values. The solid lines are the experimental results of the probability $P$ of whether to mask the input. The dotted lines are the experimental results of the probability $p$ of Bernoulli distribution.}\label{fig:select-pandP}
\end{figure}

\begin{table*}[t]
\centering
\scalebox{0.87}{
\begin{tabular}{@{}llllllllll@{}}
\toprule
Source($\downarrow$) & Attack & VGG-16 & ResNet-50 & MoNet-V2 & AdvEff-b0 & ViT-B/16 & DeiT-B & ResMLP-36 & SwinMLP-B \\ \midrule
 & PGD & 5.4 & 2.6 & 6.7 & 0.6 & 9.6 & 7.6 & 18.5 & 4.5 \\
 & PGD+ours & 27.7{\color[HTML]{32CB00} (+22.3)} & 14.6{\color[HTML]{32CB00} (+12.0)} & 30.7{\color[HTML]{32CB00} (+24.0)} & 4.7{\color[HTML]{32CB00} (+4.1)} & 11.1{\color[HTML]{32CB00} (+1.5)} & 13.8{\color[HTML]{32CB00} (+6.2)} & 33.7{\color[HTML]{32CB00} (+15.2)} & 12.5{\color[HTML]{32CB00} (+8.0)} \\
 & MIM & 21.3 & 9.60 & 23.6 & 4.7 & 8.7 & 12.3 & 21.5 & 9.1 \\
 & MIM+ours & 41.6{\color[HTML]{32CB00} (+20.3)} & 22.6{\color[HTML]{32CB00} (+13.0)} & 43.9{\color[HTML]{32CB00} (+20.3)} & 8.8{\color[HTML]{32CB00} (+4.1)} & 22.0{\color[HTML]{32CB00} (+13.3)} & 38.7{\color[HTML]{32CB00} (+26.4)} & 52.2{\color[HTML]{32CB00} (+30.7)} & 20.7{\color[HTML]{32CB00} (+11.6)} \\
 & DIM & 8.9 & 6.1 & 13.9 & 2.1 & 7.1 & 7.7 & 26.1 & 12.7 \\
 & DIM+ours & 31.4{\color[HTML]{32CB00} (+22.5)} & 18.0{\color[HTML]{32CB00} (+11.9)} & 36.6{\color[HTML]{32CB00} (+22.7)} & 6.0{\color[HTML]{32CB00} (+3.9)} & 16.7{\color[HTML]{32CB00} (+9.6)} & 22.5{\color[HTML]{32CB00} (+14.8)} & 57.8{\color[HTML]{32CB00} (+31.7)} & 25.2{\color[HTML]{32CB00} (+12.5)} \\
 & TIM & 5.6 & 2.5 & 6.8 & 0.8 & 2.8 & 2.3 & 9.2 & 4.4 \\
\multirow{-8}{*}{Mixer-B/16} & TIM+ours & 27.5{\color[HTML]{32CB00} (+21.9)} & 13.8{\color[HTML]{32CB00} (+11.3)} & 30.6{\color[HTML]{32CB00} (+23.8)} & 4.1{\color[HTML]{32CB00} (+3.3)} & 11.5{\color[HTML]{32CB00} (+8.7)} & 15.2{\color[HTML]{32CB00} (+12.9)} & 33.8{\color[HTML]{32CB00} (+24.6)} & 12.9{\color[HTML]{32CB00} (+8.5)} \\ \midrule
 & PGD & 3.4 & 1.5 & 4.3 & 0.6 & 1.3 & 0.3 & 2.5 & 1.5 \\
 & PGD+ours & 10.8{\color[HTML]{32CB00} (+7.4)} & 5.0{\color[HTML]{32CB00} (+3.5)} & 12.6{\color[HTML]{32CB00} (+8.3)} & 1.3{\color[HTML]{32CB00} (+0.7)} & 2.1{\color[HTML]{32CB00} (+0.8)} & 1.8{\color[HTML]{32CB00} (+1.5)} & 7.6{\color[HTML]{32CB00} (+5.1)} & 2.4{\color[HTML]{32CB00} (+0.9)} \\
 & MIM & 16.9 & 7.0 & 16.4 & 2.3 & 4.9 & 2.8 & 8.5 & 3.9 \\
 & MIM+ours & 26.5{\color[HTML]{32CB00} (+9.6)} & 12.4{\color[HTML]{32CB00} (+5.4)} & 26.9{\color[HTML]{32CB00} (+10.5)} & 4.1{\color[HTML]{32CB00} (+1.8)} & 6.2{\color[HTML]{32CB00} (+1.3)} & 5.9{\color[HTML]{32CB00} (+3.1)} & 13.6{\color[HTML]{32CB00} (+5.1)} & 3.8(-0.1) \\
 & DIM & 5.4 & 3.7 & 7.4 & 0.7 & 2.5 & 2.1 & 6.9 & 3.4 \\
 & DIM+ours & 18.8{\color[HTML]{32CB00} (+13.4)} & 11.4{\color[HTML]{32CB00} (+7.7)} & 25.8{\color[HTML]{32CB00} (+18.4)} & 3.3{\color[HTML]{32CB00} (+2.6)} & 11.8{\color[HTML]{32CB00} (+9.3)} & 15.1{\color[HTML]{32CB00} (+13.0)} & 44.9{\color[HTML]{32CB00} (+38.0)} & 14.2{\color[HTML]{32CB00} (+10.8)} \\
 & TIM & 3.3 & 2.0 & 4.4 & 0.6 & 1.2 & 0.4 & 2.9 & 1.4 \\
\multirow{-8}{*}{Mixer-L/16} & TIM+ours & 10.9{\color[HTML]{32CB00} (+7.6)} & 4.9{\color[HTML]{32CB00} (+2.9)} & 12.6{\color[HTML]{32CB00} (+8.2)} & 1.2{\color[HTML]{32CB00} (+0.6)} & 3.0{\color[HTML]{32CB00} (+1.8)} & 1.6{\color[HTML]{32CB00} (+1.2)} & 7.8{\color[HTML]{32CB00} (+4.9)} & 2.4{\color[HTML]{32CB00} (+1.0)} \\ \midrule
 & PGD & 10.1 & 4.3 & 11.8 & 2.3 & 6.3 & 13.2 & 30.9 & 5.4 \\
 & PGD+ours & 14.1{\color[HTML]{32CB00} (+4.0)} & 8.3{\color[HTML]{32CB00} (+4.0)} & 17.5{\color[HTML]{32CB00} (+5.7)} & 2.9{\color[HTML]{32CB00} (+0.6)} & 9.5{\color[HTML]{32CB00} (+3.2)} & 25.4{\color[HTML]{32CB00} (+12.2)} & 55.3{\color[HTML]{32CB00} (+24.4)} & 11.7{\color[HTML]{32CB00} (+6.3)} \\
 & MIM & 32.6 & 15.3 & 31.2 & 6.7 & 15.5 & 23.8 & 46.9 & 12.9 \\
 & MIM+ours & 33.1{\color[HTML]{32CB00} (+0.5)} & 19.3{\color[HTML]{32CB00} (+4.0)} & 35.1{\color[HTML]{32CB00} (+3.9)} & 8.8{\color[HTML]{32CB00} (+2.1)} & 22.9{\color[HTML]{32CB00} (+7.4)} & 39.0{\color[HTML]{32CB00} (+15.2)} & 71.6{\color[HTML]{32CB00} (+24.7)} & 20.0{\color[HTML]{32CB00} (+7.1)} \\
 & DIM & 24.1 & 19.2 & 33.0 & 7.3 & 22.6 & 40.7 & 73.0 & 23.8 \\
 & DIM+ours & 26.6{\color[HTML]{32CB00} (+2.5)} & 20.2{\color[HTML]{32CB00} (+1.0)} & 35.2{\color[HTML]{32CB00} (+2.2)} & 7.9{\color[HTML]{32CB00} (+0.6)} & 24.8{\color[HTML]{32CB00} (+2.2)} & 45.1{\color[HTML]{32CB00} (+4.4)} & 79.5{\color[HTML]{32CB00} (+6.5)} & 28.1{\color[HTML]{32CB00} (+4.3)} \\
 & TIM & 9.2 & 4.4 & 12.5 & 1.8 & 5.1 & 12.9 & 32.4 & 4.9 \\
\multirow{-8}{*}{Mixer-Glu} & TIM+ours & 12.6{\color[HTML]{32CB00} (+3.4)} & 6.9{\color[HTML]{32CB00} (+2.5)} & 17.1{\color[HTML]{32CB00} (+4.6)} & 2.4{\color[HTML]{32CB00} (+0.6)} & 10.4{\color[HTML]{32CB00} (+5.3)} & 24.2{\color[HTML]{32CB00} (+11.3)} & 56.8{\color[HTML]{32CB00} (+24.4)} & 10.4{\color[HTML]{32CB00} (+5.5)} \\ \bottomrule
\end{tabular}}
\caption{The fooling rate (\%) on 1k ImageNet val. The adversarial examples at $\epsilon \leq $ 16. The adversarial examples generated by our proposed MA method has a significantly higher fooling rate. We generated adversarial examples on Mixer-B/16,  Mixer-L/16 and Mixer-Glu, and conducted transferability experiments on networks of different architectures. We report the fooling rate of adversarial examples against VGG-16, ResNet-50, MobileNet-V2 (MoNet-V2), AdvEfficientNet-b0 (AdvEff-b0), ViT-B/16, DeiT-B, ResMLP-36 and SwinMLP-B}\label{table:results}
\end{table*}

\begin{figure*}[t]
  \centering
  \subfigure[Mixer-B/16 with PGD]{
    \includegraphics[width=0.23\linewidth]{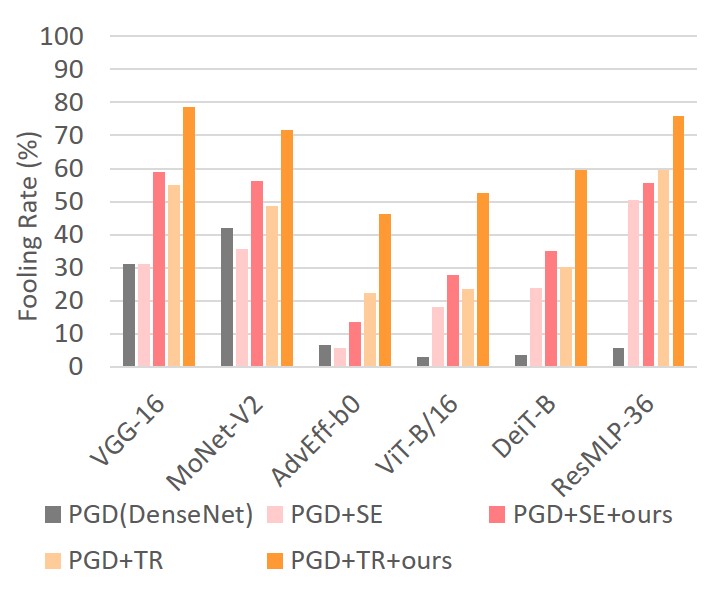}
    \label{fig:b16-a}
  }
  \subfigure[Mixer-B/16 with MIM]{
    \includegraphics[width=0.23\linewidth]{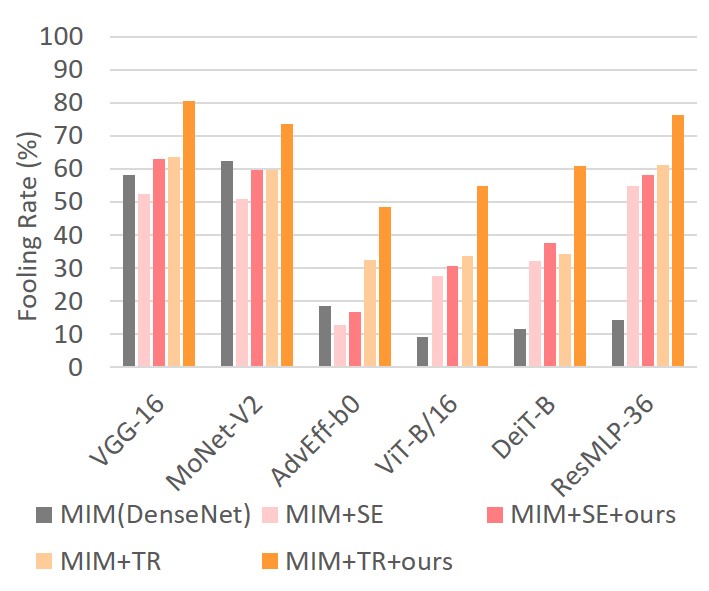}
    \label{fig:b16-b}
  }
  \subfigure[Mixer-B/16 with DIM]{
    \includegraphics[width=0.23\linewidth]{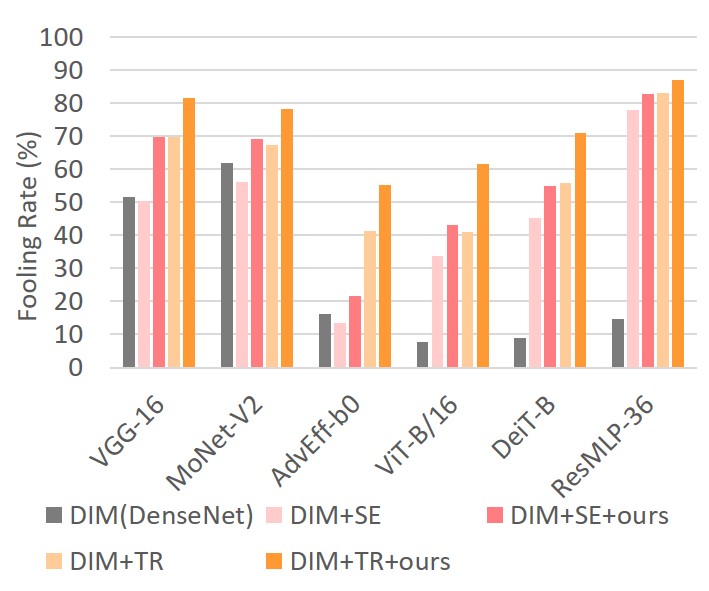}
    \label{fig:b16-c}
  }
  \subfigure[Mixer-B/16 with TIM]{
    \includegraphics[width=0.23\linewidth]{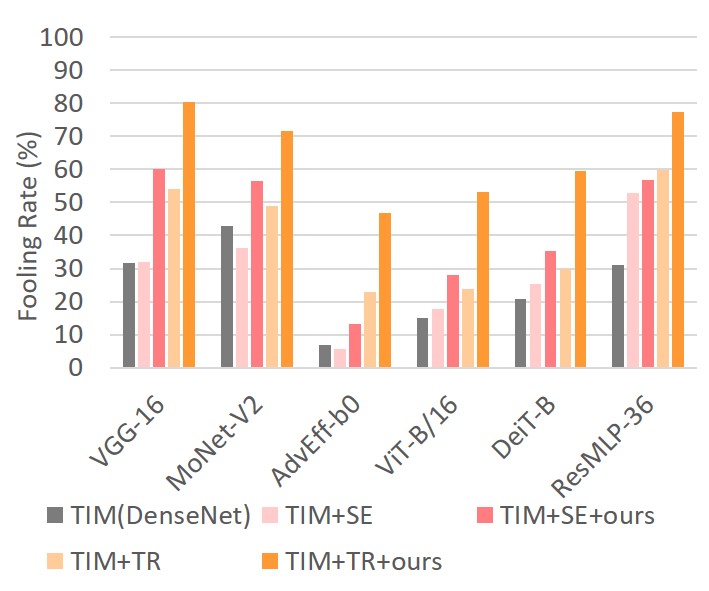}
    \label{fig:b16-d}
  }
  \subfigure[Mixer-L/16 with PGD]{
    \includegraphics[width=0.23\linewidth]{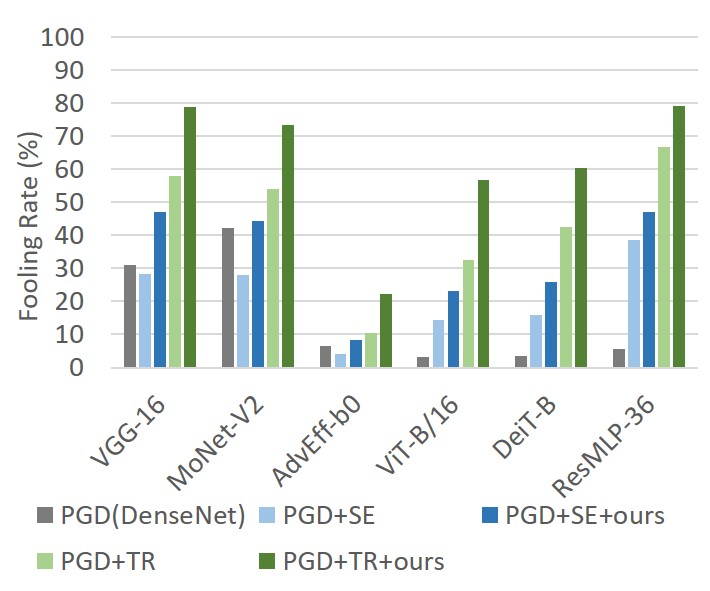}
    \label{fig:l16-a}
  }
  \subfigure[Mixer-L/16 with MIM]{
    \includegraphics[width=0.23\linewidth]{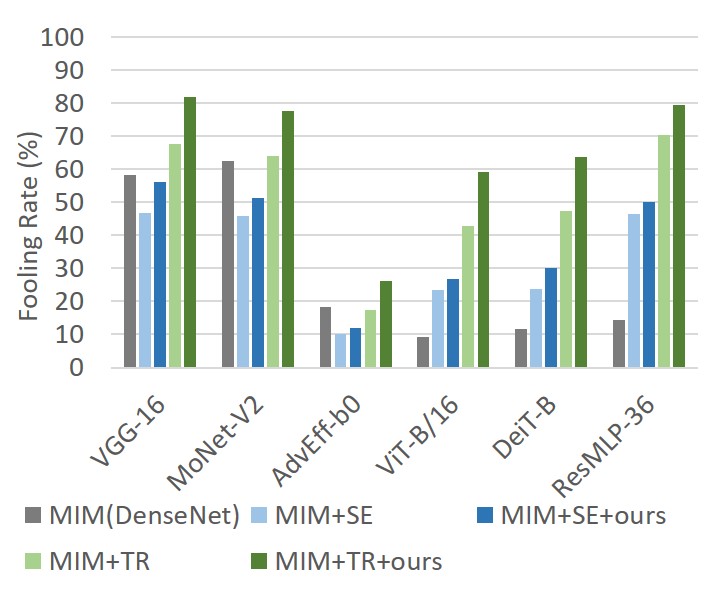}
    \label{fig:l16-b}
  }
  \subfigure[Mixer-L/16 with DIM]{
    \includegraphics[width=0.23\linewidth]{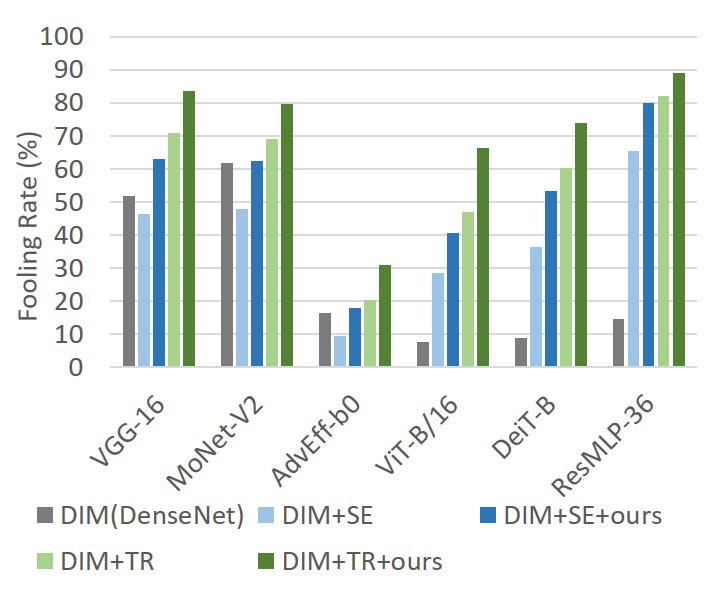}
    \label{fig:l16-c}
  }
  \subfigure[Mixer-L/16 with TIM]{
    \includegraphics[width=0.23\linewidth]{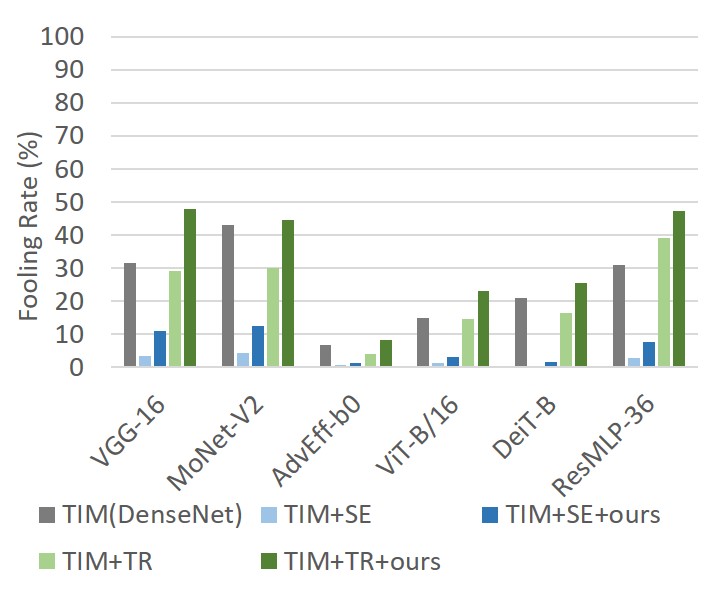}
    \label{fig:l16-d}
  }
  \caption{The fooling rate (\%) on 1k ImageNet val. The adversarial examples at $\epsilon \leq $ 16. The fooling rate of the original SE and TR methods combined with PGD, MIM, DIM, and TIM, and the fooling rate of these methods combined with our method. The source model in the first row is Mixer-B/16, and the source model in the second row is Mixer-L/16. The gray bar represents the source model is DenseNet-201. The target models are VGG-16, MobileNet-V2 (MoNet-V2), AdvEfficientNet-b0 (AdvEff-b0), ViT-B/16, DeiT-B and ResMLP-36.}
  \label{fig:b16andl16}
\end{figure*}

\begin{figure}[t]
    \centering
    \includegraphics[width=0.85\linewidth]{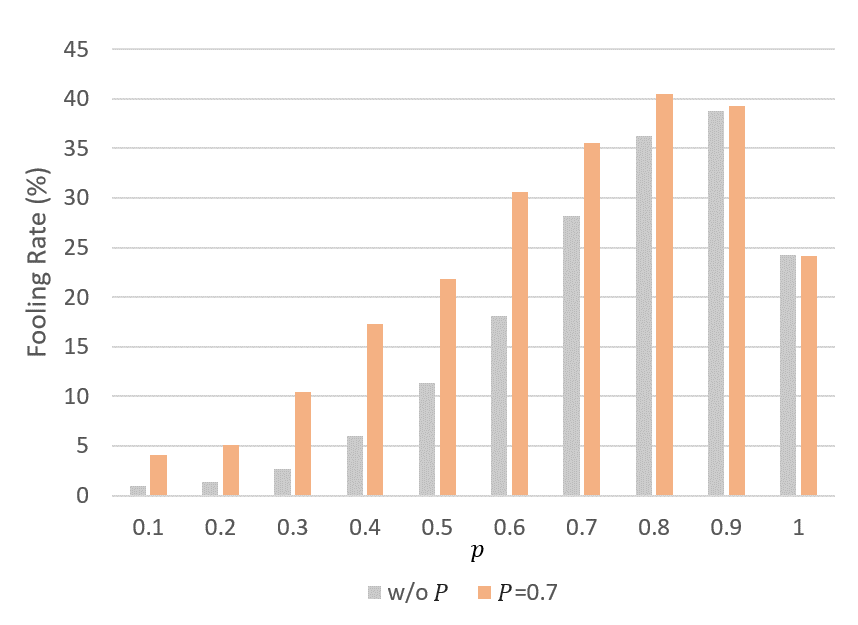}
    \caption{Ablation experiment. Compare the experimental results of deleting the probability $P$ of whether to mask the input and setting it to 0.7. The horizontal axis is the value of the probability $p$ of the Bernoulli distribution.}\label{fig:ablative}
\end{figure}

\section{Experiments}
\subsection{Settings}
We choose Mixer-B/16, Mixer-L/16, and Mixer-Glu with GluMLP \cite{shazeer2020glu} in MLP-Mixer as source models. We also select the classic model DenseNet201 in CNNs as the source model to compare with MLP-Mixer. For the target model, we report experimental results on the following models, VGG \cite{simonyan2014very} (VGG-13, VGG-16, VGG-19), ResNet \cite{he2016deep} (Resnet-18, Resnet-34, Resnet-50, Resnet-101, Resnet-152), DenseNet (Densenet-121, Densenet-161, Densenet-169, Densenet-201), ReXNet (ReXNetV1-10, ReXNetV1-13, ReXNetV1-15, ReXNetV1-20, ReXNetV1-30) and MobileNet-V2 based on the CNN architecture, EfficientNet (AdvEfficientNet-b0, AdvEfficientNet-b1, AdvEfficientNet-b2) after adversarial training, ViT-B/16, T2T-ViT (T2T-12, T2T-14, T2T-19) and DeiT-B based on the transformer architecture, ResMLP-36 \cite{touvron2021resmlp} and SwinMLP-B \cite{liu2021swin} based on the MLP architecture. These models are provided by TIMM \cite{rw2019timm}, the experimental results of more models can be found in the supplementary material. We randomly selected 1k samples from the ImageNet validation set, and these samples can be correctly classified by all the above models. We use the fooling rate to assess the transferability of adversarial examples, i.e. the percentage of adversarial examples whose predicted labels on the target model are inconsistent with ground-truth labels. We uniformly set the perturbation budget $\epsilon$ to 16, and the number of attack iterations $T$ to 50.

\subsection{Probability values}
As shown in Eq. 3, there are two probability values that need to be set in our method, one is the probability $P$ of whether to mask the input and the other is the probability $p$ of generating the masking matrix using the Bernoulli distribution. We report the mean fooling rate of adjusted probability on each kind of models, the source model is Mixer-B/16, and the attack method is a combination of MIM, TR and our MA method. We first test the probability $P$ of whether to mask the input. We randomly set the probability $p$ of the Bernoulli distribution to 0.8. As shown by the solid line in Fig. \ref{fig:select-pandP}, when $P$ is 0.7, the fooling rate of generated adversarial examples reaches the maximum value on multiple models. Then we test the probability $p$ of the Bernoulli distribution, setting the probability $P$ to 0.7. As shown by the dotted line in Fig. \ref{fig:select-pandP}, the fooling rate of the generated adversarial examples on multiple models first rises and then declines. When $p$ is 0.8, the fooling rate reaches the maximum value. So we set $P$ to 0.7 and $p$ to 0.8.

\subsection{Improve transferability to black-box MLP-based models}

In this section, we discuss the experimental results of adversarial transferability between MLP-Mixer and black-box MLP-based models. As shown in the last two columns of Tab. \ref{table:results}, we report the experimental results of the MLP-based ResMLP-36 and SwinMLP-B as target models. For the basic adversarial attack methods PGD, MIM, DIM and TIM, combined with our method, the adversarial examples generated by Mixer-B/16 can improve the fooling rate on ResMLP-36 by about 20\%. After DIM combined with our method, Mixer-L/16 generated adversarial examples can improve the fooling rate by 38.0\% on ResMLP-36. As shown in Fig. \ref{fig:b16andl16}, the SE and TR methods combined with our method can further improve the transferability of adversarial examples on ResMLP-36. Experimental results demonstrate that our method is able to break the channel-mixing and token-mixing mechanisms of MLP-Mixer by masking the input of each Mixer layer, and improve the transferability of adversarial examples on the MLP-based models.

\subsection{Improve  transferability to black-box ViTs}
As shown in Tab. \ref{table:results}, we report the experimental results of transformer-based ViT-B/16 and Deit-B as target models. After MIM combined with our method, Mixer-B/16 generated adversarial examples can improve the fooling rate by 26.4\% on DeiT-B. As shown in Fig. \ref{fig:b16andl16}, we observe that although the SE and TR algorithms are proposed for ViTs, the transferability of MLP-Mixer generated adversarial examples is further improved on the transformer-based models when combined with our MA method. The experimental results demonstrate that our method can achieve a Dropout-like effect by masking the input of each Mixer layer, preventing adversarial examples from overfitting the source model MLP-Mixer, and improving the transferability of adversarial examples to non-MLP models.

\subsection{Improve  transferability to black-box CNNs}

As shown in Tab. \ref{table:results}, we report the experimental results of CNN-based VGG-16, ResNet-50, MobileNet-V2 and AdvEfficientNet-b0 as target models. PGD, MIM, DIM and TIM, when combined with our method, can improve the fooling rate of adversarial examples generated by Mixer-B/16 on VGG-16 by more than 20\%. The results on the AdvEfficientNet-b0 model after adversarial training demonstrate that our attack method is not only effective for ordinary CNNs, but also for robust models.

It is worth noting that, as shown in Fig. \ref{fig:b16andl16}, the SE and TR method combined with our method improves the transferability by about 20\% compared to the original method on VGG-16, MobileNet-V2 and AdvEfficient-b0. After SE and TR combined with our method, the adversarial examples generated by Mixer-B/16 are even more transferable on VGG-16, MobileNet-V2 and AdvEfficient-b0 than those generated by CNN-based DenseNet-201. As shown in Fig. \ref{fig:gradcam}, we show adversarial examples generated on MLP-B16 and GradCam \cite{jacobgilpytorchcam} images generated on VGG-16, our method can further force the target model to focus on the wrong regions in the adversarial examples compared to the original method. This fully demonstrates the effectiveness and great potential of our method.

The experimental results demonstrate that our MA method combined with existing adversarial attack methods can comprehensively improve the transferability of adversarial examples on CNN-based models, transformer-based models and MLP-based models. This means that, using only our method, generating adversarial examples on a single model transfers well on CNN-based models, transformer-based models and MLP-based models. Our method achieves the effect of using an ensemble model on a single model, but uses fewer resources and is faster.

\subsection{Ablation Study}
We perform ablation experiments on our method. The attack method is a combination of MIM, TR and our method. The source model is Mixer-B/16, and the target model DenseNet-201. The experimental results are shown in Fig. \ref{fig:ablative}. Deleting the probability $P$ of whether to mask the channel, the fooling rate is decreased compared to setting $P$ to 0.7. Although the fooling rate is flat after the probability $p$ of Bernoulli distribution is greater than 0.9, it does not exceed the maximum value at $P$=0.7, which proves that every part of our method contributes.

\section{Conclusion}
We propose a novel transfer-based attack, called Maxwell’s demon Attack (MA). By using MA to mask the part input of each Mixer layer of the MLP-Mixer, we are able to greatly improve the transferability of its generated adversarial examples. On some CNN-based models, the adversarial examples generated by our method on the MLP-Mixer even exceed the transferability of the adversarial examples generated using CNNs. Our method can simply be combined with existing adversarial attack methods against CNNs and ViTs. We conduct experiments on models with multiple architectures, and the experimental results demonstrate the superiority of our method. To our knowledge, we are the first work to study the transferability of MLP-Mixer.

\bibliographystyle{named}
\bibliography{ijcai22}

\end{document}